# FORECASTING THE JSE TOP 40 USING LONG SHORT-TERM MEMORY NETWORKS

Adam Balusik, Jared de Magalhaes, Rendani Mbuvha[1]

School Statistics and Actuarial Science, University of the Witwatersrand, Johannesburg, 2050,
South Africa
abalusik@gmail.com , jared.dem@gmail.com , Rendani.mbuvha@wits.ac.za

**Abstract**

As a result of the greater availability of big data, as well as the decreasing costs and increasing power of modern computing, the use of artificial neural networks for financial time series forecasting is once again a major topic of discussion and research in the financial world. Despite this academic focus, there are still contrasting opinions and bodies of literature on which artificial neural networks perform the best and whether or not they outperform the forecasting capabilities of conventional time series models. This paper uses a long-short term memory network to perform financial time series forecasting on the return data of the JSE Top 40 index. Furthermore, the forecasting performance of the long-short term memory network is compared to the forecasting performance of a seasonal autoregressive integrated moving average model. This paper evaluates the varying approaches presented in the existing literature and ultimately, compares the authors' approach and results to that existing literature. The paper concludes that the long short-term memory network outperforms the seasonal autoregressive integrated moving average model when forecasting intraday directional movements as well as when forecasting the index close price.

## 1. INTRODUCTION

Traditionally, time series forecasting was the predominantly employed method of financial market analysis and forecasting (Samarawickrama & Fernando, 2017). However, following recent economic and financial market crashes, a disconnect between the financial market and the economy has emerged and their performances are not as highly correlated as traditional thinking depicts (Schöneburg, 1990). This disconnect, and the financial markets' volatility and noise characteristics have resulted in share price forecasting becoming one of the most difficult tasks among time series forecasting (Wang, Huang & Wang, 2012). Furthermore, investors are required to identify and act on trading opportunities quicker than ever due to the improvement in market efficiency and communication speeds (Schöneburg, 1990). Therefore, traditional methods of forecasting, such as the more basic methods of fundamental and technical analysis as well as more complex time series modelling, are becoming outdated and lack the forecasting accuracy they once had. Evidently, there is an increasing need for a more accurate method of financial market forecasting. This research attempts to address this problem by implementing a form of a neural network in order to forecast the financial market returns.

[1] Corresponding author

The purpose of this research is to evaluate the potential of using long short-term memory (LSTM) recurrent neural network (RNN) models, hereon referred to as a LSTM network, for forecasting intraday returns of the JSE Top 40 index. Section 4.4 explains the network in greater detail, but it is essentially a model that attempts to use time series data as well as internally processed versions of said data to determine relationships in the data which can be used to forecast future values of the time series. This potential is discussed in comparison to the traditional time series group of forecasting methods.

The paper is organised as follows. Section 2 states the aim of the research and discusses its importance, practical relevance, and contribution to knowledge. Section 3 then relates this aim to past literature in the form of a literature review. Section 4 outlines what a LSTM is, the data that will be used, the methodology imposed in fitting the respective LSTM and time series models and how the results will be evaluated. Section 5 is an analysis and comparison of the results obtained from the respective models. Lastly, section 6 summarises the findings of the paper before drawing up a final conclusion.

## 2. RESEARCH QUESTIONS AND CONTRIBUTIONS

The central aim of this work is to develop a LSTM network to be used to forecast, using historical data, and the intraday movements of the JSE Top 40 index. In doing so, three main questions are to be answered:

Can a LSTM network be used to effectively forecast the direction of movement of the JSE Top 40 index?

- Can a LSTM network effectively forecast the intraday returns of the JSE Top 40 index?
- Are LSTM networks more effective than traditional time series models in forecasting the JSE top 40 index?

This research is set against the backdrop of a world that is becoming faster paced due to the rise in computer technology. It is estimated that 90% of trading that occurs happens with some form of assistance by computers[2]. Thus, in terms of this research, the reason for its importance is that of relevance. To compete in this increasingly computerised world, investors have to deal with large amounts of data. Hansson (2017) speaks of the effective use of LSTM networks for processing this data, particularly when compared to standard time series models. Their efficacy lying in their use of massive computing power to find connections within data, which would otherwise be overlooked if traditional methods of analysis are used. As mentioned in the introduction, investors have a need for a more accurate financial forecasting method.

LSTM networks have been applied in various international markets and indices such as the Dow Jones Industrial index in the United States of America (Troiano, Villa, & Loia, 2018), the Shanghai Composite index (Zhuge, Xu, & Zhang, 2009) and the Swedish OMX index (Hansson, 2017). LSTM networks have yet to be applied widely in the South African market, particularly to the JSE Top 40 index. In its application, this paper hopes to provide starting literature for the subject of using recurrent neural networks in financial forecasting in South Africa. A LSTM network is used, rather than a general RNN, as it can incorporate possible lagged dependencies occurring through time (Zhuge, Xu, & Zhang, 2009). The use of LSTM

[2]Dani Burger, The U.S. Stock Market Belongs to Bots, 2017. https://www.cnbc.com/2017/06/13/death-of-the-human-investor-just-10-percent-of-trading-is-regular-stock-picking-jpmorgan-estimates.html 22 March 2018

networks in time series analysis is a new and developing field especially when compared traditional time series analysis; as such it offers a new approach to forecasting financial markets in South Africa.

The modelling framework used can also easily be extended to forecast movements in other areas of the financial market as well (e.g. single share value forecasting and forecasting derivative prices) (Samarawickrama & Fernando, 2017). The research also allows for various and different types of inputs to be used, and for possible new correlations and relationships between data to be found.

## 3. RELATED WORK

Various studies have been performed to evaluate the performance of LSTM networks in financial market forecasting as well as how their performance compares to that of traditional time series models. However, as discussed in the aim, LSTM networks have not yet been applied in a South African context to the JSE Top 40 index. Therefore, this literature review evaluates past literature on the use of artificial neural networks (ANNs), and more specifically LSTM networks, for forecasting financial markets in general. The literature is also analysed to determine the appropriateness of using a LSTM network, as opposed to another possible neural network, for the purpose of forecasting financial market movements.

### 3.1 Is a LSTM Network the Most Appropriate Neural Network to Use for Forecasting?

Lawrence (1997) explains how various approaches such as fundamental, technical and time series analysis have been used in an attempt to forecast share prices and compares these approaches to the novel use of neural networks in share price forecasting. However, the relative merits of different neural networks are not considered by Lawrence (1997). Therefore, the relative merits of two of the main classes of neural networks, namely Multi-layered Perceptron (MLP) networks, and RNNs are discussed below and finally the appropriateness of a LSTM network is considered.

An ANN is a neural network designed to attempt to replicate the functioning of the human brain. This is done through rationally making decisions by determining the non-linear relationships in the input data (Zhang, Patuwo & Hu, 1998). The main advantages of an ANN are as follows:

- ANNs are data driven and as such do not need a priori assumptions about the data (Zhang, Patuwo & Hu, 1998).
- ANNs can generalise and infer relationships in a data set after being trained using a sample of the data (Zhang, Patuwo & Hu, 1998).
- ANNs are non-linear (Zhang, Patuwo & Hu, 1998). Section 3.4 expands on this point explaining how it adds to an ANNs effectiveness over time series models.

The above points add credibility to the notion of using ANNs to forecast financial share price movements.

A MLP network is one of the simplest forms of an ANN. Figure 1 below, shows a basic MLP network. When being trained, the MLP network outputs forecasted values by passing the inputs through hidden layers that determine the relative importance of the inputs and subsequently weight them accordingly. The forecasted output is compared with the actual output and the error is used to repeat the above process and adjust the weights in order to minimise this error. This repeated process is known as backpropagation (BP in figure 1). Huang (2009) describes how the use of backpropagation in an MLP network helps minimise the errors between the outputs forecast by the network and the actual outputs from the data. However, the shortcomings of a MLP network are the inability to process outputs back into input nodes, since it is a feed forward network, and the need for large input nodes which increase computing time (Huang, 2009). Smith, Beyers & De Villiers (2016) use a feed forward MLP network for share price forecasting and note that other neural network structures, such as a RNN, are likely to be more effective since they overcome the feed forward shortcoming discussed above.

**Figure 1**
Basic (3 – 4 – 3 -1) MLP Network

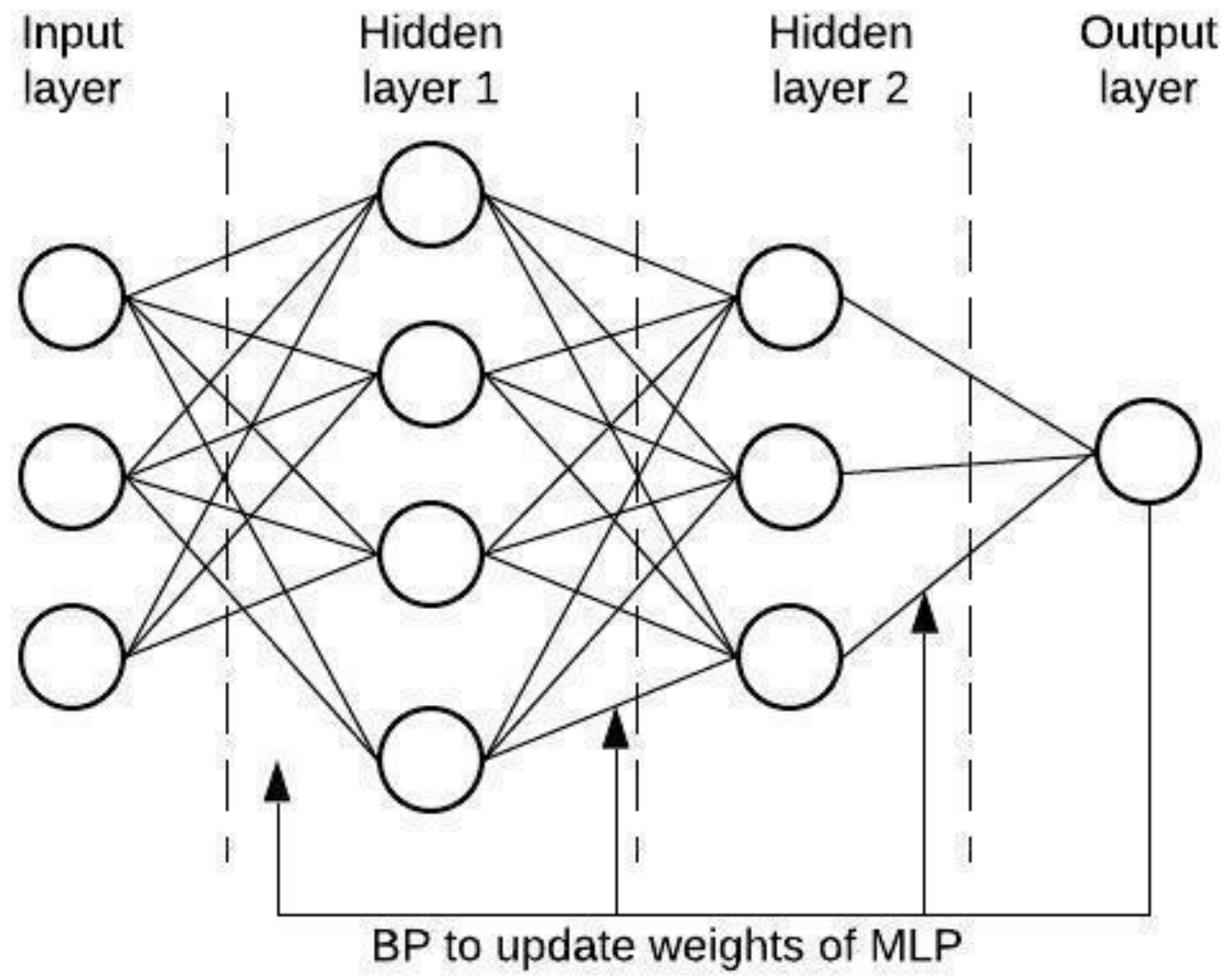

A merit of a RNN is that it addresses one of the shortcomings of the MLP network in that it not only makes use of backpropagation but it can also take the output from the output nodes and feed the values back into the network (Samarawickrama & Fernando, 2017). Pascau, Mikolov, & Bengio (2013) explain how this property makes RNNs useful for finding correlations between data that are separated in time, which is useful in our case because stock prices may depend on underlying variables which change over time. However, this also makes the model difficult to train due to the exploding gradient problem and vanishing gradient problem (Pascau, Mikolov, & Bengio, 2013). This is a problem where when using conventional backpropagation through time, the errors used to update the nodes of the network will either exponentially increase, exploding gradient problem, or exponentially diminish, vanishing gradient problem, depending on the size of the weights (Hochreiter & Schmidhuber, 1997). In essence, this means that a RNN cannot effectively use long-past historical data to make future predictions.

LSTM networks provide a solution to the exploding gradient problem and vanishing gradient problem. The inner-workings of a LSTM network are explained in section 4.4, however the novelty of the approach is explained here. Proposed in Hochreiter & Schmidhuber (1997), a LSTM network uses multiplicative units to allow the creation of a memory cell which helps the backpropagation to determine what information is relevant and what information is not. This is why a LSTM network is proposed as a method to forecast financial market returns, it can use long short-term memory cells to remember long-term dependencies and then use these to more effectively forecast share prices.

### 3.2 Can LSTM Networks Effectively Forecast the Directional Movement of Financial Market Returns?

Hansson (2017) evaluated an LSTM network and found that it resulted in significant forecasting of the directional movements for the Swedish OMX index; however, the same could not be concluded for the S&P 500 or Bovespa index. In contrast to this, Di Persio & Honchar (2016) developed an LSTM network that effectively forecast the directional movements of the S&P 500, as well as the Forex EUR/USD returns. The reasoning behind the contradictory results regarding the S&P 500 is unclear and Hansson (2017) states that it is difficult to understand why the LSTM network used did not successfully forecast the S&P 500 because of the large hidden parameter space of the neural network. Out of five different neural networks that Di Persio & Honchar (2016) proposed, the LSTM network performed the second best behind a wavelet convolution neural network.

Overall, the results of past literature regarding the effectiveness of LSTM networks in forecasting the directional movement of financial market returns is promising and showed better results than forecasting returns in general.

### 3.3 Can LSTMs Effectively Forecast Financial Market Returns?

Hansson (2017) found no concrete evidence to support the effective forecasting of financial market returns using LSTM networks. He argues that the forecasted movements of the network do not significantly follow the actual movements seen in the markers observed. This result was in contradiction to three other studies.

Firstly, Samarawickrama & Fernando (2017) concluded that LSTM networks performed well when forecasting financial market returns and generally produced lower forecasting errors than other recurrent neural networks. Samarawickrama & Fernando (2017) also stated that out of the 40 proposed RNNs, it was one of the LSTM networks that had the lowest forecasting error and was determined to be the best RNN. However, Samarawickrama & Fernando (2017) determined that overall the multilayer perceptron group of models performed the best. However, they explicitly stated that this was due to only using the past two days' worth of data as inputs and if this was increased, LSTM networks would perform the best.

Secondly, when considering the Shanghai Composite index, Zhuge, Xu & Zhang (2017) concluded that their LSTM model displayed excellent forecasting performance and that the successful forecasting of the opening price of the share was due to the ability of the LSTM network to learn the long-term dependence of the time series data. This is a promising finding for LSTM networks in general as this is a characteristic they all display.

Lastly, Bao, Yue & Rao (2017) developed a complex LSTM network that also consisted of wavelet transforms and stacked autoencoders. Bao, Yue & Rao (2017) produced results that showed that this complex model could forecast financial market returns very accurately and with a high level of significance, but simpler LSTM networks still showed some potential in forecasting financial market returns.

Lawrence (1997) comments on the idea that in the existence of an efficient market, that the efficient market hypothesis (EMH) would imply that a neural network, or other forecasting methods such as traditional time series analysis, should not be able to predict future movements. (Hansson, 2017) finds that LSTM networks may be useful in efficient markets where the weak form of the EMH does not exist.

Overall, the results of past literature regarding the effectiveness of LSTM networks in forecasting financial market returns is promising but is highly dependent on both the specific LSTM network design (Samarawickrama & Fernando, 2017) and the financial market in question (Lawrence, 1997).

### 3.4 Are LSTM Networks More Effective than Traditional Time Series Models in Financial Market Forecasting?

The seasonal component of time series data may be better captured by time series models, such as the ARIMA model (Smith, Beyers & De Villiers, 2016). However, the use of time series models results in smoothing of the data which may lead to small fluctuations in share prices becoming unrecognisable (Schöneburg, 1990). This is unwanted as modern investors are shifting their focus to profiting from small price changes on a large number of shares (Schöneburg, 1990). Compared to time series models, ANNs, in general, are more applicable to real-world problems because they can make use of the non-linear relationships in data, whereas the former cannot (Zhang, Patuwo & Hu, 1998). Zhuge, Xu & Zhang (2017) state that this benefit is especially apparent in LSTMs due to their ability to learn the long-term dependence of the time series data. Furthermore, ANNs perform well when dealing with incomplete and inaccurate data and unclear problems, which are often the cases in reality (Korol, 2013).

Smith, Beyers & De Villiers (2016) found that even the simplest ANNs forecast simple share prices as accurately as time series models. When looking at LSTMs in particular, Hansson (2017) found that they showed similar output results to those of traditional time series models for financial market forecasting but outperformed the time series models when considering the directional movements of financial market returns. Hansson (2017) concluded that LSTM networks produce equal or better forecasting results when compared to even the most advanced time series models.

Overall, the majority of past literature agrees that LSTM networks are more effective than traditional time series models in financial market forecasting.

## 4. METHODS

### 4.1 Data Used

For input data, daily closing prices of the JSE Top 40 index are used as well as closing values of the SSE 100 index are used. Closing data is used as input data into the LSTM network. This is the simplest data to collect and is used in multiple LSTM models such as in Zhuge, Xu & Zhang (2017) and Di Persio & Honchar (2016). The SSE 100 index is a composite index of the top 100 shares that are traded on the Shanghai Stock Exchange. The SSE index is used as a proxy of the strong trade relationship that South Africa and China exhibit. Angomoko (2017) describes how South Africa’s exports to China have increased by 10% from 1992 to 2012. Along with South Africa’s involvement in BRICS (a trade partnership between Brazil, Russia, India, China and South Africa), the Chinese economy has grown in size and become one of the largest economies in the world with large exports (He & Zhang ,2010) and there has been recent investment in South Africa by China in mid-2018 (the Chinese government agreeing to R196 billion of investment and loans provided to Eskom and Transnet, two state-owned enterprises)[3]. These factors would indicate that the Chinese market should have some sway in the movements of the South African economy and hence the JSE top 40.

The daily data used is collected from IRESS Expert (formerly McGregor BFA) spanning 2 years from 15 December 2015 to 15 December 2017. This yields 501 data entries. The data will be split into a training set and a testing set. The training data set is set for 12 December 2015 to 11 August 2017 to 29 January 2017(yielding 412 values, just over 80% of the data). The remainder is used as the testing set (14 August 2017 to 15 December 2017).

Since the indices operate in different markets, not all values exist for corresponding dates. The dates used are based off of available data for the closing prices of the JSE top 40 index. To ensure there are corresponding data values for the SSE index, in the presence of a missing value the data value for the most recent date is used.

Figure 2 shows the scaled values of the closing prices for the JSE top 40 index and the SSE 100 index. The values are scaled to allow for better and easier comparison between the datasets.
**Figure 2**

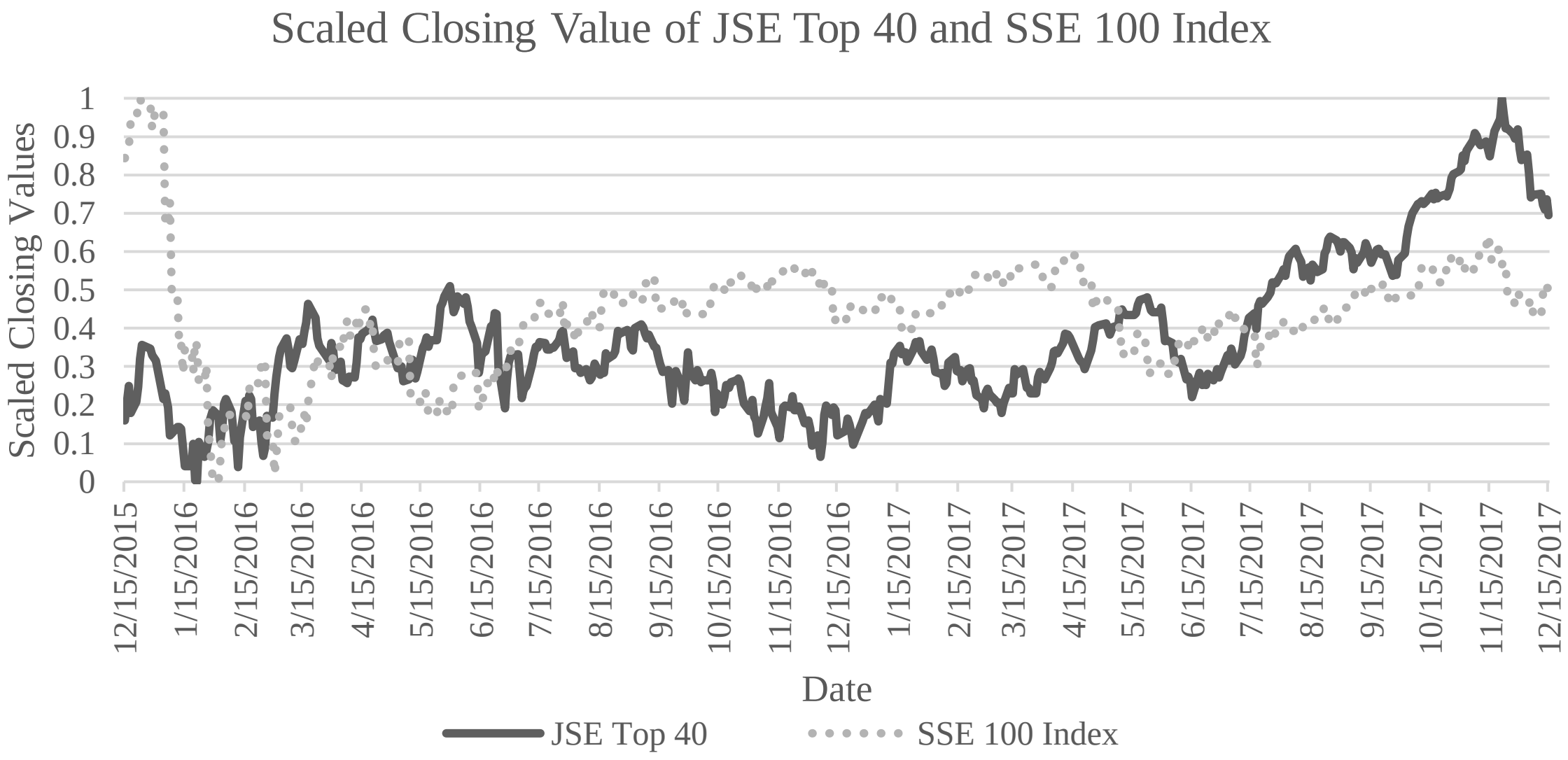

## 4.2 High Level Overview of Method

The input data mentioned in paragraph 4.2.2 is used as input into the LSTM network. The LSTM network is coded in Python and uses Keras with a TensorFlow backend. Closing values for the JSE top 40 index are output (these are the forecasted values).

An appropriate auto-regressive integrated moving average (ARIMA) time series model, described in section 4.4, is determined and fitted to the data. The data is then forecasted for the same time period as that done for the LSTM network.

---

[3] Luke Daniel, Three major agreements just signed by China and South Africa, 24 July 2017. https://www.thesouthafrican.com/three-major-agreements-just-signed-by-china-and-south-africa/ 07 October 2018

The forecasted movements for the LSTM network and the time series model are compare on three criteria, namely: ability to forecast up and down movements of the index in the following day, overall predictive performance (measured using the root mean-squared error(RMSE), and the accuracy to forecast overall return. These analyses are described in section 4.6

## 4.3 Time Series Methodology

The data used for the time series analysis is as described above in section 4.2 with closing prices from 12 December 2015 to 11 August 2017 being used to fit the time series model and closing prices from 14 August 2017 to 15 December 2017 being used to test the accuracy of the model forecasts. The time series model fitting and forecasting is done using the R software and its forecasting package.

A seasonal autoregressive integrated moving average (SARIMA) time series model is selected due to its ability to account for the seasonality present in the time series, the autocorrelation of the time series itself (autoregressive component), the autocorrelation of the error terms (moving average component) and the trend that is observed in the data (integrated component).

In general, the interpretation of the parameters of an SARIMA $(p, d, q)x(P,D,Q)_s$ is as follows. The time series needs to be differenced d times in order to remove the trend and the value of the time series at some time t is modelled using p of its previous values (the autoregressive component) and the error terms at time t, t-1, … , t-q (the moving average component). Similarly, P and Q are the orders of the autoregressive and moving average seasonal component of the SARIMA model, while D is the order of seasonal differencing. Lastly, the s represents the number of seasonal periods per year for the time series.

The SARIMA model is fitted manually using the Box and Jenkins Methodology and determines that the best fitting model is SARIMA $(2, 0, 2)x(0,1,0)_{250}$. The reason for the number of seasonal periods being 250 is due to there, on average, being 250 days a year where the financial markets are open and hence there are 250 closing prices for the index each year. A standardised residuals plot is checked to ensure that the residuals are random about zero, which is a requirement of the model. Furthermore, the autocorrelation function of the residuals is plotted at different lags to ensure that the autocorrelation has been accounted for. Finally, the Ljung-Box statistic is plotted at different lags to ensure that there is no significance at any of them. Once the three tests above are all passed it is concluded that the model fits well.

The SARIMA model is fitted automatically using the built in R function and the optimal model is again SARIMA $(2, 0, 2)x(0,1,0)_{250}$. Using the SARIMA $(2, 0, 2)x(0,1,0)_{250}$ model closing prices for testing period, of 89 days, are forecast. The MSE of the forecasts is then calculated and compared to that of the LSTM network in order to evaluate their respective forecasting accuracy.

## 4.4 General Workings of a LSTM Network

LSTM networks are special cases of RNNs (they use feedback loops to inform their decision making by processing input recurrently through the hidden layers several times). The term “memory” denotes the fact that a LSTM network can work with lags of arbitrary length and duration by having states which allow its cells to store values implying that it should either ‘forget’ or ‘remember’ information, this allows it to remember information for long periods of time. Hochreiter & Schmidhuber (1997) originally developed the network as a solution to the vanishing gradient problem. Similarly explained in Zhuge, Xu, & Zhang (2009) and Hansson (2017), Figure 3, below, shows the inner workings of a cell in a LSTM network.

**Figure 3**

Cell in a LSTM Network

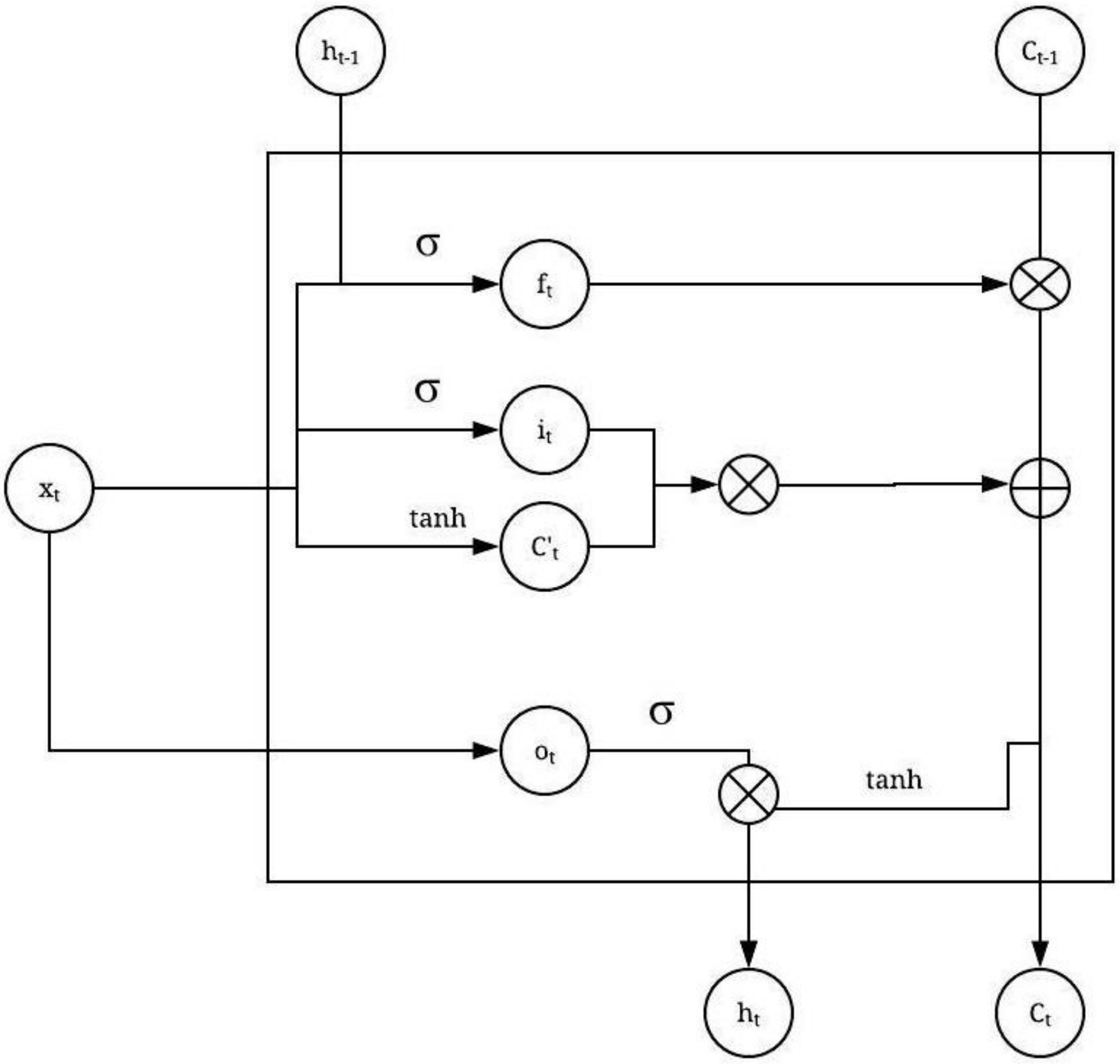

t: the time.

$x_t$: the input into the cell.

$C_t$: this is the internal state of the cell; it is what allows for 'memory' (both short-term and long-term) to be stored.

$h_t$: the value of the hidden layer, it is the output of the cell at time t. This is the value that will be used in the next cell of the network.

$f_t$: the "forget gate", it is a sigmoid layer(σ) which looks at the input into the cell as well at the previous hidden layer( $h_{t-1}$) and determines how much of this information(represented as a number between 0 and 1) to add into the cell state $C_{t-1}$. Note a 0 would indicate all information is forgotten, while a 1 would indicate all information being retained.

$i_t$: the "input gate", it is a sigmoid layer(σ) which determines how much of the input information should enter the cell state.

$C'_t$: the input modulation gate. Since the input gate is a sigmoid layer, it will output values between 0 and 1 and, hence, never allow for information to be forgotten. The input modulation gate uses a tanh layer to allow the cell to forget memory.

$o_t$: the "output gate". It is a sigmoid layer ($\sigma$) that determines, with input from the cell state, how much of the information should be retained for the next cell and output ($h_t$).

Below are the equations for the gates and layers, note that W_ and b_ represent a weight and a bias, with respect to the gate that is referenced in their subscripts:

$$\sigma(x) = \frac{1}{1+e^{-x}}$$

$$\tanh(x) = \frac{e^{x} - e^{-x}}{e^{x} + e^{-x}}$$

$$f_t = \sigma ( W_f \cdot [ h_{t-1} , x_t ] + b_f )$$

$$i_t = \sigma ( W_i \cdot [ h_{t-1} , x_t ] + b_i )$$

$$C'_t = \tanh ( W_C \cdot [ h_{t-1} , x_t ] + b_C )$$

$$C_t = f_t . C_{t-1} + i_t . C'_t$$

$$o_t = \sigma (W_o \cdot [ h_{t-1} , x_t ] )+ b_o$$

$$h_t = o_t . \tanh ( C_t\, t )$$

### 4.5 LSTM Network Methodology

The data used for the training and testing of the network is as described above in section 4.2 with data from 12 December 2015 to 11 August 2017 being used as input to train the network. Closing prices are output and compared to those of the data from 14 August 2017 to 15 December 2017, this will be used to test the network. The data is normalised and rescaled to be between 0 and 1 using the MinMaxScaler function, this is done to make the optimisation process faster.

The training data is passed through the network and is used to determine the weights and biases that should be used by the network to make its forecasted. Once trained the network is then tested, using the testing set, to determine its accuracy.

The network is coded in Python and uses the Keras package and TensorFlow as its backend. This is done as the output of the network is the main focus of the paper, as opposed to the structure of the network.

As mentioned in section 4.3, since the main focus of the paper is the output of the network and not the structure of the LSTM network, a relatively simple network is used. The model is built sequentially meaning that each layer only has connections to nodes in layers directly before or after it. Tanh activation functions are used inside the layers. In compiling, mean absolute error (MAE) is used as the loss function and Adam is used as the optimisation algorithm. The Adam (adaptive moment estimation) algorithm is an alternative to the classical stochastic gradient descent procedure used to update network weights. It is used as it combined the benefits from other algorithms. Namely, the AdaGrad (Adaptive Gradient Algorithm) and its ability to maintain a learning rate for each network weight, and the RMSProp (Root Mean Square

Propagation) algorithm which adapts these maintained network weight specific learning rates on the basis of how quickly they change. This algorithm is often used for its efficiency.

Following this the model is fitted to the data. 200 epochs are used to train the data. As described in Hansson (2017). An epoch being a single pass in training a neural network in which the training data is fed into the model and the network's weights are updated. A batch size of 200 is used. The batch size limits the number of samples to be shown to a network before it updates weights, i.e. it is the division of the training data into batches (Hansson, 2017).

The network outputs a forecast of the closing prices for the dates in the test period. These closing prices are still scaled and thus are unscaled to provide the forecasted closing values for the index. The testing set is used as a benchmark or performance measurement for the network as it is the actual observed values of the network. A measure of the accuracy of the model is using the RMSE, described in section 4.6.

### 4.6 Analysis of Results

The network's ability to forecast if the movements of the index will be up or down in the following day. The accuracy of this is easily tested by assigning a 1 for every time the network is correct and a 0 for every time it is incorrect. An overall proportion greater than 50% will indicate that the network can forecast movements better than pure chance. This is also compared to the proportion generated by the time series model.

The overall predictive performance of the LSTM network. This can be assessed through calculating the root mean-squared error (RMSE). This RMSE approach is similar to the one followed by Zhuge, Xu, & Zhang (2017). In the paper the mean-squared error is used, here RMSE is used to allow for easier comparison between the respective RMSE values of the LSTM and the time series model. RMSE is calculated as follows:

$$\text{RMSE}=\sqrt{\frac{\sum_{t=1}^{n}\left(\text{actual(t)-forecast(t)}\right)^2}{n}}$$

, where actual(t) denotes the original observation of the index at time t and forecast(t) is the forecasted value by the network at time t. For the LSTM network and the time series model case, n is given as 498 (This is the number of forecasted values).The smaller the value of the MSE, the more accurate the network is at forecasting the index. A MSE for the time series model is also calculated and compared to that of the LSTM network

The accuracy of the network to forecast the return over the time period is measured by calculating the total return over the period. This is calculated as follows:

$$\frac{\text{forecasted closing }(t = N)\text{ - forecasted closing }(t = 0)}{\text{forecasted closing }(t = 0)}$$

, where "forecasted closing (t = N)"is the forecasted closing value of the index at the last time point of the data set (i.e. the most recent time point which is 15 December 2017) and "forecasted closing(t=0)"is the forecasted closing value of the index at the first time point of the data set. The total return over the period forecasted by the time series is also calculated. Both of these values calculated are compared to the actual total return of the index over the time considered.

Another measure of the accuracy of the network and the time series model to forecast movements is the generated average daily rate of return based on forecasted values. It is calculated as follows:

$(1 + i_{avg})^n = \prod_{t=1}^{n} (1+ i_t)$ , where

$$i_t = \frac{\text{forecasted closing (t) - forecasted closing (t - 1)}}{\text{forecasted closing (t - 1)}}$$

Where:

$i_t$ is the daily return calculated

$i_{avg}$ is the equivalent average daily rate of return for the period

n is the number of days used in the test data set

forecasted closing (t) is the forecasted closing price of the JSE top 40 index at time t

$i_{avg}$ is calculated for both the time series model and the LSTM using the values that they forecast. These values are compared to the actual $i_{avg}$ over the period (it is calculated in the same way, except the actual values of the JSE Top 40 index are used in place of the forecasted values). The calculated average daily rates of return are compared. The closer a predictive model's average daily rate of return to the actual value, the better the model is at predicting the movements of the index.

## 5. RESULTS AND DISCUSSION

### 5.1 Forecasted Values

The LSTM network's generated forecast values are shown graphically and plotted against the actual JSE Top 40 values; this is shown in Figure 4 which can be found in the Appendix.

The forecasted values generated by the time series model values are shown graphically and plotted against the actual JSE Top 40 values; this is shown in Figure 5 which can be found in the Appendix.

The LSTM network forecast moves in line with the index data. Furthermore, it captures some aspect of the volatility of the close prices around the long-term trend which is observed by the LSTM network forecast displaying similar peaks and troughs to that of the index data. On the contrary, the time series model only partially captures some of the peaks and troughs of the index data. It does not, however, reflect the slight upward trend of the index. Instead it incorrectly forecasts values representative of a downward market.

### 5.2 Forecasting Up/Down Movements

**Table 1**

Accuracy in Forecasting the Direction of Intraday Movements of the JSE Top 40 Index

| Method | Predictive Accuracy |
|---|---|
| LSTM Network | 0,500 |
| Time Series Model | 0,443 |

The table shows that the LSTM Network accurately predicts the direction of the movement of the index 50% of the time while the time series model accurately forecasted the direction of movement 44% of the time.

The LSTM network is slightly better at forecasting the directional movements of the index. This is in agreement with Hanson (2017) where it was concluded that the LSTM network performed better than the time series model used.

It should further be noted that the LSTM network better forecasts the peaks and troughs of the index. This is beneficial for short-term forecasting, which is imperative when building a trading strategy.

The lack of significant predictive accuracy of either model mean that the models should not be used to try and predict daily directional movements. Instead, the LSTM is more useful in predicting longer term trends.

## 5.3 Root Mean-Squared Error

**Table 2**
RMSE of Model Forecasts

| Method | RMSE |
|---|---|
| LSTM Network | 372,2169 |
| Time Series Model | 6628,2240 |

The significantly lower RMSE value for the LSTM network compared to that of the time series model is indicative that the LSTM network approach generates more meaningful insight into the shape of the data. This follows the finding in Samarawickrama & Fernando (2017) that an LSTM can be an effective model to forecast share performance, better than that of time series models.

## 5.4 Forecasted Overall Return

**Table 3**
Returns Forecast by Models for Overall Period 14 August 2017 to 15 December 2017

| Method | Forecast Overall Return |
|---|---|
| LSTM Network | 5,4321% |
| Time Series Model | -0,0988% |

The actual return of the index over the same period was calculated to be 3,7563%. Both the time series model and the LSTM network under predicted the overall return of the index. However, the time series model inaccurately predicted a negative return.

The difference between forecasted overall return and actual overall return gives some insight into the overall predictive power of the models used. The LSTM being more effective than the time series model.

## 5.5 Forecasted Average Daily Return

**Table 4**

Average Daily Returns Forecast by Models over Period 14 August 2017 to 15 December 2017

| Method | Forecast Average Daily |
|---|---|
| LSTM Network | 0,000 601 |
| Time Series Model | -0,001 18 |

The actual average daily return of the index over the same period was calculated to be 0,000 419. The values are used to represent the average returns per day over the aforementioned period.

Comparing forecasted to actual average daily returns instead of the overall return shows how the LSTM slightly over predicts daily movements while the time series model under predicts the daily movements. This is also reflected in the graphs of the forecasted daily movements.

# 6. SUMMARY AND CONCLUSION

There are various options for the types of neural network that can be implemented when attempting to forecast financial time series data and a fair body of literature is in agreement that LSTM networks are one of the better options. However, all previous literature has used international indices and not those of South Africa. Therefore, the authors implement a LSTM network in order to forecast the closing prices and intraday directional movements of the JSE Top 40 index and compare its performance to that of a traditional ARIMA time series model. When considering the intraday directional movement forecasts, the ARIMA model underperformed relative to the LSTM network, there is still much room for improvement for the LSTM to better predict directional movements. When considering the closing price forecasts, the LSTM network performed significantly better than the ARIMA model. Furthermore, the LSTM network fitted the index data significantly better than the time series model.

The inability of either method to fully forecast the movements of the index (particularly the daily directional movements) can be related to the strong form of the efficient market hypothesis (EMH). That is to say that the South African equity market is efficient in its pricing of information in the index. This agrees with the commentary of Hansson (2017) on how the efficiency of the market is likely to reduce the significance in meaningful index forecasting in markets.

The outperformance of the LSTM relative to the time series model shows the promise of LSTMs for predicting share prices. A more complex LSTM network structure could be used to try better predictive performance. Multiple factors could be tested to see if the performance of the network could be improved. Things such as number of hidden layers, activation functions, optimisation algorithms, batch size, and number of epochs could all be varied and tried to determine their efficacy on predictive performance. Another extension could be to use additional inputs (possible inputs could be technical data or macro-economic data) in an attempt to explain more of the directional movements of the index.

However, with these extensions, there is the possibility of adding too much to the LSTM. This not only complicates the network but it may result in the network overfitting the model to the data. This can result in the network's predictive power decreasing (Lawrence, 1997).

A more advanced time series model, namely a combination of ARIMA and generalized autoregressive conditional heteroscedasticity (GARCH) processes, may better fit the non-linear movement present in the data. This combination will allow the autoregressive nature of the time series model to be captured by the ARIMA model, while the non-symmetric conditional movements of the variance will be captured by the GARCH model. Both methods will likely benefit from running multiple tests to generate a range of predictions that could be used to create a confidence interval of the forecasted results.

ACKNOWLEDGEMENTS
This paper acknowledges the supervision provided by Mr R Mbuvha throughout the writing of the paper. Mr P Castilla is also acknowledged for the use of his work as the base for the LSTM network structure used in this paper. His code and work can be found at https://www.kaggle.com/pablocastilla/predict-stock-prices-with-lstm .

**APPENDIX**

**Figure 4**

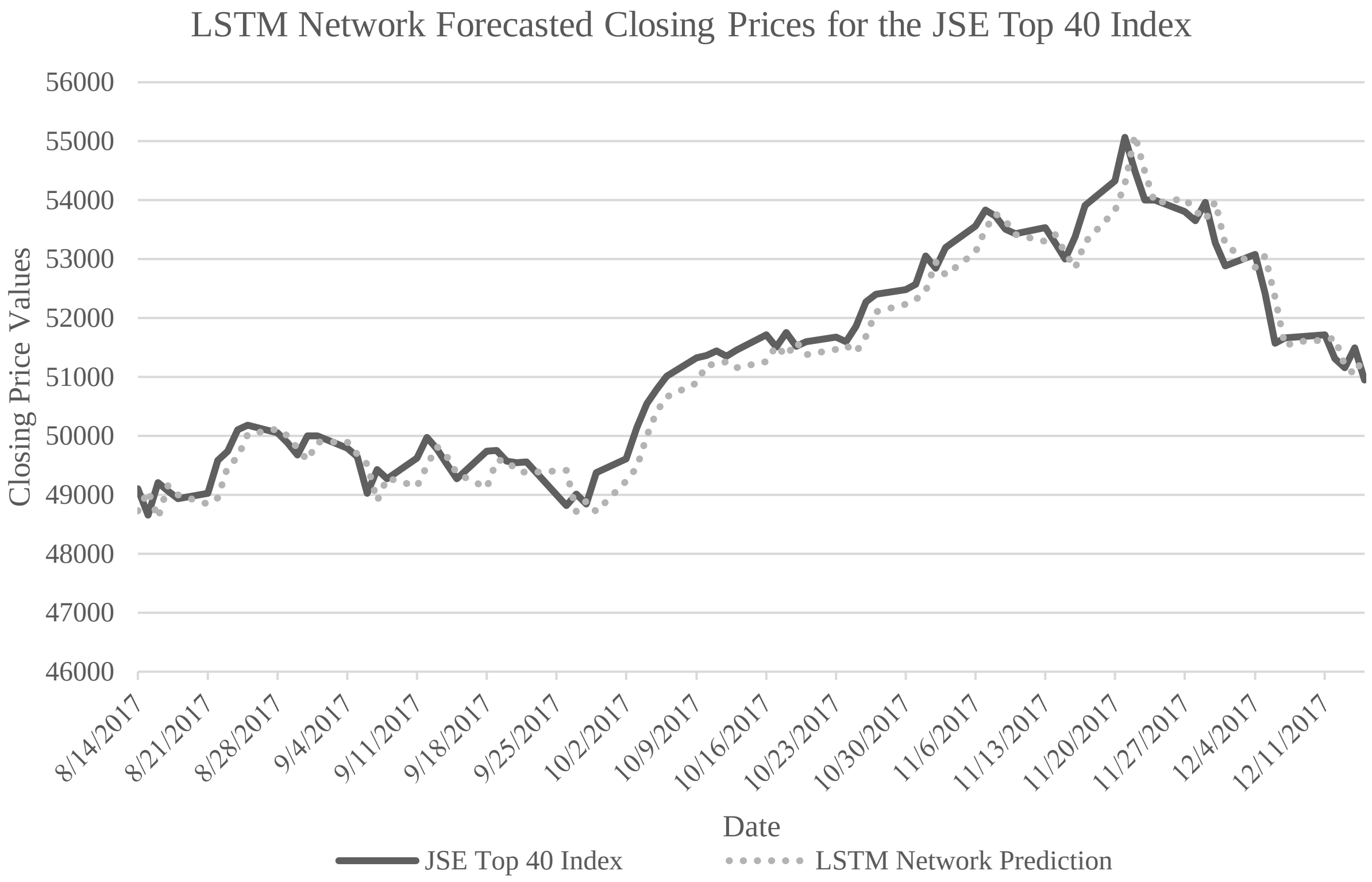

**Figure 5**

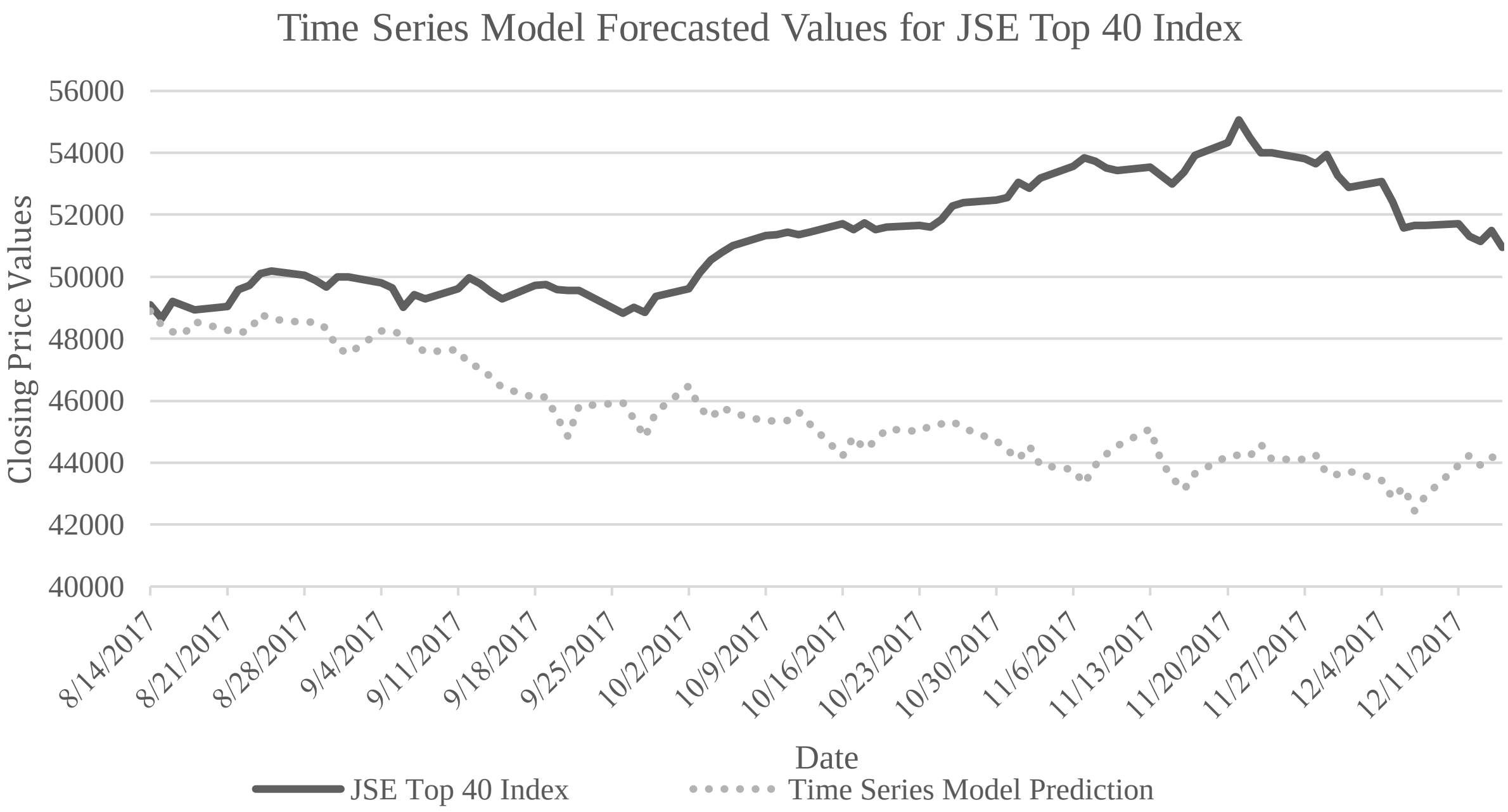